\documentclass{article}
\usepackage[letterpaper, margin=1.5in]{geometry}

\usepackage{setspace}
\usepackage{lineno}
\usepackage{color}
\usepackage[utf8]{inputenc}
\usepackage[hyphens]{url}  
\usepackage{graphicx} 
\usepackage{amsmath}
\usepackage{authblk}
\usepackage{hyperref}

\usepackage{multirow, multicol}
\usepackage{makecell}
\usepackage[numbers]{natbib}

\usepackage{fontawesome5}
\newcommand{\cirbd}{\mathrel{\text{\faDotCircle[regular]}}}
\newcommand{\norm}[1]{\left\lVert#1\right\rVert}

\newif\ifshortCV

\makeatletter
\ifshortCV
\let\cite\@gobble
\let\bibliography\@gobble
\fi
\makeatother

\title{Q-RBSA: High-Resolution 3D EBSD Map Generation Using An Efficient Quaternion Transformer Network}

\author[1, Cor]{Devendra K. Jangid}
\author[2]{Neal R. Brodnik}
\author[3]{McLean P. Echlin}
\author[3]{Tresa M. Pollock}
\author[2]{Samantha H. Daly}
\author[1]{B.S. Manjunath}

\affil[1]{Electrical and Computer Engineering}
\affil[2]{Mechanical Engineering}
\affil[3]{Materials Department}
\affil[Cor]{dkjangid@ucsb.edu}

\date{}

\begin{document}

\maketitle

\begin{abstract}
Gathering 3D material microstructural information is time-consuming, expensive, and energy-intensive. Acquisition of 3D data has been accelerated by developments in serial sectioning instrument capabilities; however, for crystallographic information, the electron backscatter diffraction (EBSD) imaging modality remains rate limiting. We propose a physics-based efficient deep learning framework to reduce the time and cost of collecting 3D EBSD maps. Our framework uses a quaternion residual block self-attention network (QRBSA) to generate high-resolution 3D EBSD maps from sparsely sectioned EBSD maps. In QRBSA, quaternion-valued convolution effectively learns local relations in orientation space, while self-attention in the quaternion domain captures long-range correlations. We apply our framework to 3D data collected from commercially relevant titanium alloys, showing both qualitatively and quantitatively that our method can predict missing samples (EBSD information between sparsely sectioned mapping points) as compared to high-resolution ground truth 3D EBSD maps. 
\end{abstract}

\section{Introduction}
In the pursuit of the development of materials for extreme environments, 3D microstructural information is essential input for structure-property models \cite{ICME2008}. Many engineering materials are polycrystalline, meaning they are composed of many smaller crystals called grains, and the arrangement of these grains impacts their thermomechanical properties. To collect crystallographic microstructure information, 3D microscopy techniques have been developed that span lengthscales from nanoscale to mesoscale \cite{EchlinCOSSMS2020}. These experiments require costly or challenging to access equipment, like synchrotron light sources for high X-ray fluxes \cite{Miller2020,Bernier2020,Reischig_COSSMS_2020}, precise automated robotic mechanical polishing and imaging \cite{RowenhorstCOSSMS2020,Chapman2021}, or high-energy ion beams and/or short pulse lasers coupled to electron microscopes \cite{Echlin2021,Garner2022}. Recent advancements in 3D experimentation have greatly reduced the time required for other aspects of data collection, but serial sectioning methods (where material is progressively removed from the sample between images) predominately rely on electron backscatter diffraction (EBSD) imaging for collecting crystallographic information, which is three orders of magnitude slower per pixel (ms per pixel) than standard secondary electron or backscatter electron imaging ($\mu$s per pixel). 

EBSD is a scanning electron microscope (SEM) imaging modality that maps crystal lattice orientation by analyzing Kikuchi diffraction patterns that are formed when a focused electron beam is scattered by the atomic crystal structure of a material according to Bragg's law. A grid of Kikuchi patterns is collected by scanning the electron beam across the sample surface. These patterns are then indexed to form a grid of orientations, which are commonly represented as images in RGB color space using inverse pole figure (IPF) projections. EBSD maps are used to determine the microstructural properties of materials such as texture, orientation gradients, phase distributions, and point-to-point orientation correlations, all of which are critical for understanding material performance \cite{EBSD2009}. However, scaling EBSD scans to 3D by serial sectioning is a time-consuming and energy-intensive process, often requiring hundreds of millions of EBSD patterns to be collected per sample - motivating methods to reduce the number of required points, such as smart or sparse sampling \cite{Godaliyadda2018,Zhang2018,Tong2019}, or machine learning super-resolution \cite{Wang2021}. In these methods, missing information can be inferred using interpolation-based algorithms (bicubic, bilinear, or nearest neighbor) or data-based learning. Recent progress in computer vision \cite{Wang2021,dai2019san,zamir2022restormer} has shown that the generation of missing samples/data with data-based learning outperforms traditional interpolation algorithms for RGB images. However, unlike RGB images, EBSD maps have embedded crystallography, so existing learning-based methods are not well suited to generate missing EBSD data.  

In our previous work \cite{dkjangid_ebsdsr}, we developed a deep learning framework for 2D super-resolution that utilized an orientationally-aware, physics-based loss function to generate high-resolution (HR) EBSD maps from experimentally gathered low-resolution (LR) maps.  This approach allowed for significant gains in 2D resolution, but expansion to 3D remained difficult due to data availability limitations (3D EBSD is expensive and time consuming to gather). To address this, here we have designed a 3D deep learning framework based on quaternion convolution neural networks with self-attention alongside physics-based loss to super-resolve high resolution 3D maps using as little data as possible. Using real-valued convolution for quaternion-based data has been shown to be inefficient and has loss in the inter-channel relationship \cite{parcollet2018quaternion}; leading to longer training times and larger data burdens. We demonstrate that a quaternion-valued neural network is more efficient and produces better results than real-valued convolution neural networks such as those used in previous work \cite{dkjangid_ebsdsr}.

The crystallographic material information contained in EBSD maps is generally expressed in the form of crystal orientation spatially resolved at each pixel or voxel. These orientations, like other rotational data, can be expressed unambiguously using quaternions. They therefore can be incorporated into network architecture as prior information by using quaternion-valued convolution for local-level correlation, rather than real-valued or complex-valued convolution. The basic component in traditional CNN-based architectures is real-valued convolutional layers, which extract high-dimensional structural information using a set of convolution kernels. This approach is well-suited for unconstrained image data like RGB, but when convolution kernels fail to account for strict inter-channel dependencies where present, the result is greater learning complexity. Some successful efforts have been made to design lower-complexity architectures by extending real-valued convolution to complex-valued convolution \cite{trabelsi2017deep, aizenberg2018image} and quaternion-valued convolution \cite{matsui2004quaternion, kusamichi2004new, isokawa2009quaternionic} in the field of robotics \cite{yun2006design}, speech and text processing \cite{parcollet2018quaternion}, computer graphics \cite{shoemake1985animating, pletinckx1989quaternion}, and computer vision \cite{isokawa2009quaternionic, Zhu_2018_ECCV}. Although these convolution layers are useful to learn local correlations, they struggle to learn long-range correlations, whereas transformer-based architectures have recently shown significant success in learning long-range correlations in natural language \cite{vaswani2017attention} and high-level vision tasks \cite{zamir2022restormer, liang2021swinir}. However, the computational complexity of transformer-based architectures grows quadratically with the spatial resolution of input images due to self-attention layers, so transformers alone are not well-suited for restoration tasks. However, recent work by Zamir \cite{zamir2022restormer} proposed self-attention across channel dimensions to reduce complexity from quadratic to linear with progressive learning for image restorations and showed superior results to convolution-based architecture alone. 

Inspired by this idea, we propose the use of quaternion self-attention with progressive learning to incorporate long-range material relationships into super-resolved EBSD maps, while having linear computational scaling with the spatial size of the map. Progressive learning refers to having variable patch sizes instead of fixed patch sizes during training, which is relevant for most engineered material microstructures, where important features can span across length scales (and patch sizes). Titanium alloys, for example, are well-known to have many different microstructural variants accessible via processing, resulting in varying grain size and morphology. For the two alloys presented here, the Ti-6Al-4V variant has smaller equiaxed grains, while the Ti-7Al alloy has much larger grain size, so applying a fixed patch size would be sub-optimal across these two materials. To enforce long-range learning among these grain features, we used progressive patch sizes starting from 16 to 100 during the training of the network. Training behaves in a similar fashion to curriculum learning processes where the network starts with a simpler task and gradually moves to learning more complex ones. 

\begin{figure}
    \centering
    \includegraphics[width=1\linewidth]{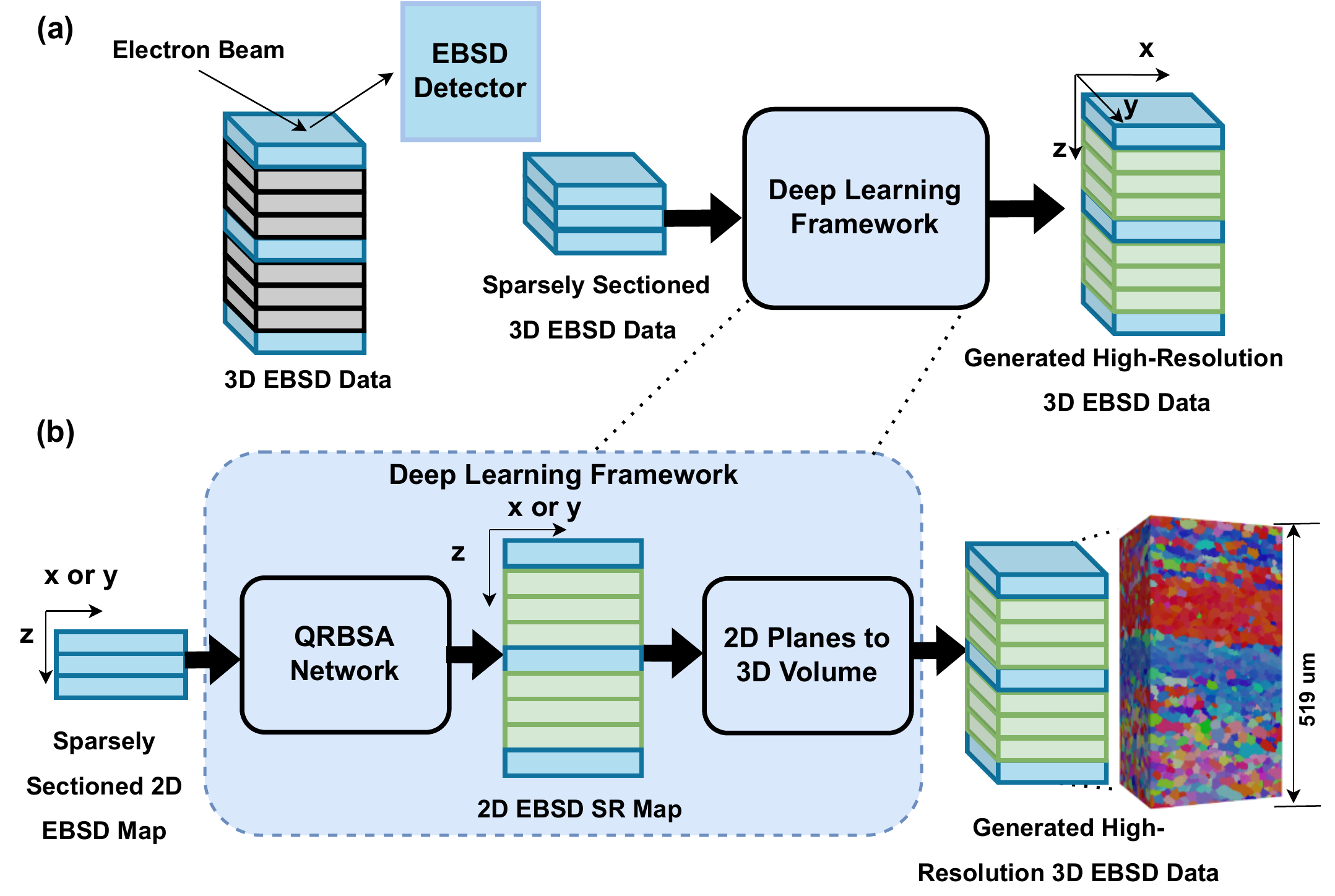}
    \caption{\textsc{Q-RBSA EBSD Resolution Enhancement Framework:} 
   In the experimental pipeline shown in (a), material researchers collect EBSD orientation information for each (x,y) coordinate in a given sectioning plane, and then remove material using laser ablation or robotic polishing to reach the next plane in the z direction to build a 3D volume. In our framework, researchers collect EBSD information from a reduced set of points (blue planes), omitting some planes that would normally be gathered (gray planes). The missing information (green planes) are then generated in 2D as a series of (x,z) or (y,z) planes by our quaternion-based, physics-informed deep learning framework, shown in (b). Here, the network takes advantage of orthogonal independence to efficiently generate 3D volumes using less data, as large amounts of EBSD are costly and the choice of serial sectioning direction has minimal impact on the resultant final volume.
    }
    \label{fig:ebsd3d}
\end{figure}

\section{Results}

\subsection{Deep Learning Framework:}
The objective of our framework is to generate missing sample planes from experimental 3D EBSD data that is sparse along the z-axis.  In this approach, material researchers collect sparsely sectioned 3D EBSD data (blue planes) as shown in figure \ref{fig:ebsd3d} (a), due to the high cost associated with serial sectioning and collecting 3D EBSD data at higher resolution. Ideally, a 3D deep learning framework would be designed to generate the missing planes (gray planes), but experimental EBSD data is costly to gather, so available 3D data is extremely limited. Additionally, 3D neural networks require more learned parameters, which, with limited available data, increases the likelihood of overfitting. Instead of a full 3D architecture, a deep learning network is implemented on 2D EBSD maps orthogonal to the sectioned planes, shown as the xz or yz planes in figure \ref{fig:ebsd3d} (b). Our network takes sparsely sectioned xz or yz EBSD maps as input to generate the missing rows normal to the z-axis. The generated 2D EBSD maps are then combined into a 3D volume. EBSD collection is a point-based scanning method that is directionally independent; therefore missing z rows can be generated from xz or yz EBSD maps, and two 3D volumes can be formed from each sparsely sectioned dataset. 

\subsection{Network Architecture:}
Although EBSD maps are visualized similarly to RGB images, they are multidimensional maps with inter-channel relationships, where crystal orientation is described using Euler angles, quaternions, matrices, or axis-angle pairs. Our previous work \cite{dkjangid_ebsdsr} demonstrated that quaternion representation is well-suited to orientation expression for loss function design, due to its efficient rotation simplification and avoidance of ambiguous representation. However, we previously used real-valued convolution layers to learn features, which is sub-optimal for EBSD orientation maps, as inter-channel quaternion relations are lost. Generally speaking, convolution networks provide local connectivity and translation equivariance, which are desirable properties for images, but if additional feature correlations are going to be learned efficiently, it is critical to encode relevant structural modalities into the network architecture and loss function.  Real-valued convolution can still learn quaternion inter-channel information, but it requires extra network complexity, and by consequence, additional data to inform that complexity. Here, the use of quaternion convolution efficiently encodes prior orientation information into kernels, and also has the advantage of reducing the number of trainable parameters by 4, as detailed in the supplement. 

\begin{figure}
    \centering
    \includegraphics[width=1\linewidth]{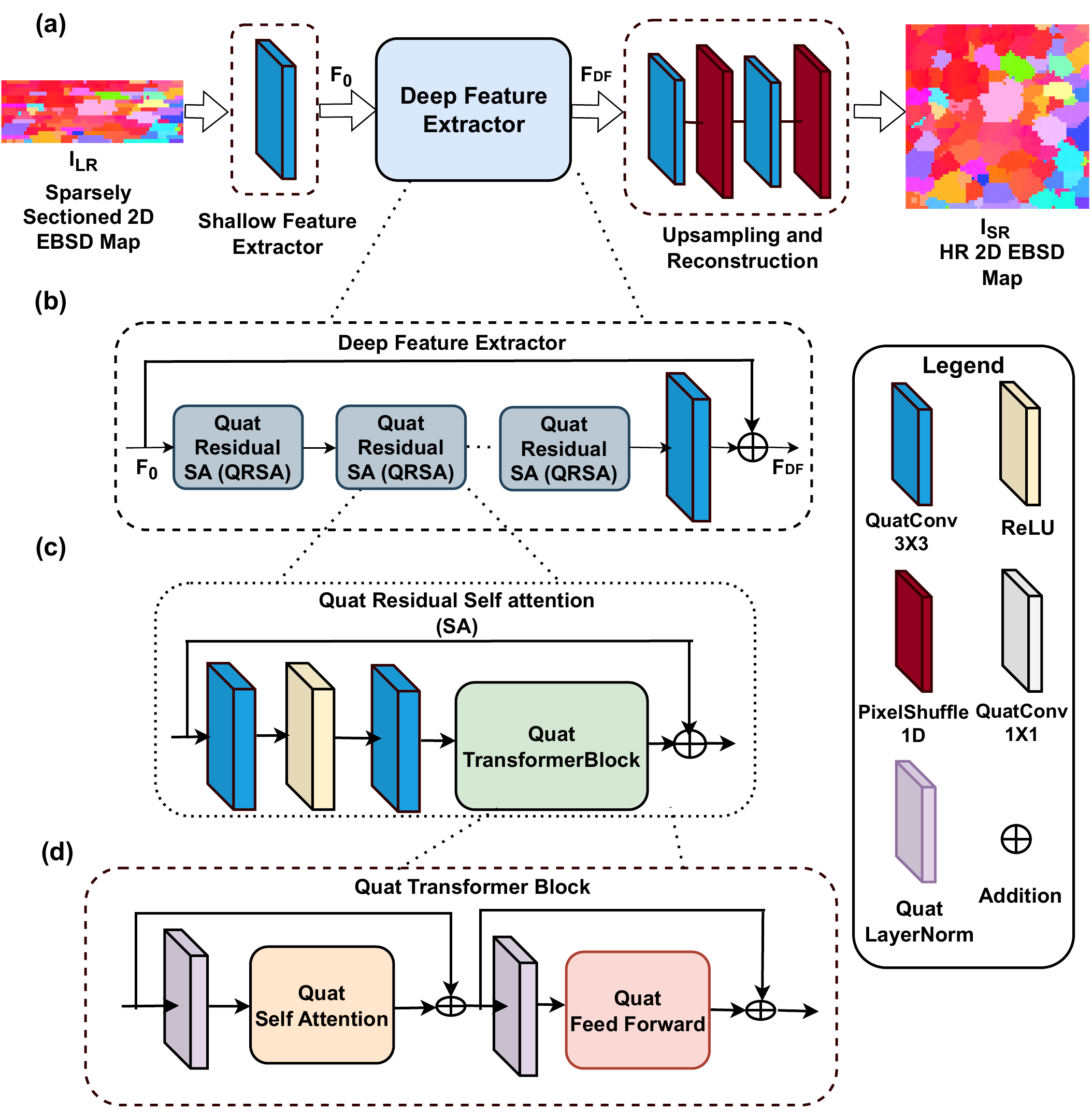}
    \caption{\textsc{Quaternion Residual Block Self-attention (QRBSA) Network:} A sparsely sectioned 2D EBSD map is given to the QRBSA network (a) to generate a high-resolution 2D EBSD map. QRBSA consists of three parts: a Shallow feature extractor, a Deep feature extractor, and Upsampling and Reconstruction. The deep feature extractor uses a residual architecture (b) where residual self-attention blocks (c) are modified with quaternion convolution layers and transformer blocks (d) to efficiently handle orientation data. Quaternion convolution is used to learn local-level relationships, while quaternion transformer blocks learn the global statistics of feature maps. Pixelshuffle layer, modified for 1-dimensional upsampling, is used in the upsampling and reconstruction block to upsample feature maps.}
    \label{fig:qrbsa}
\end{figure}

Quaternion convolution neural networks (QCNN) \cite{gaudet2018deep} use kernels which take the Hamilton product between the feature vectors rather than just computing correlations between features, as is done in real-valued convolution. In quaternion convolution, the feature output of each 2D convolution layer is split into four parts: the first part is the real component, and the remaining three parts are the complex vector component. Quaternion algebra is ensured by manipulating matrices of real numbers. In a QCNN, the convolution of a quaternion filter matrix with a quaternion vector is the Hamilton product. The Hamilton product between two quaternion vectors $p=( p_0, \vec{p})$  and $q=( q_0, \vec{q}) $ is defined as:
\begin{equation}
\
                (p_0, \vec{p}) \otimes (q_0, \vec{q}) = (p_0 q_0 - \vec{p} \cdot \vec{q}, \ p_0 \vec{q} + q_0 \vec{p} + \vec{p} \otimes \vec{q} )              
\end{equation}
Where, $\vec{p}=p_1 i + p_2 j + p_3 k$ and $\vec{q}=q_1 i + q_2 j + q_3 k$
 Motivated by \cite{zamir2022restormer}, we have designed a quaternion-based self-attention layer with a gated feed forward network in our network architecture, called QRBSA: Quaternion Residual Block Self-Attention as shown in figure \ref{fig:qrbsa}. This efficient neural network architecture generates high-resolution EBSD maps from 2D input maps that are sparsely sectioned along the z-direction. 

Introducing non-linearity through an activation function is not straightforward for quaternions, as the only functions that satisfy the Cauchy-Riemann-Fueter (CRF) equations in the quaternion domain are linear or constant \cite{parcollet2020survey}. However, locally analytic quaternion activation functions have been adapted for use in QNNs with standard backpropagation algorithms \cite{de1997local,isokawa2012quaternionic}. There are two classes of these quaternion-valued activation functions: fully quaternion-valued functions and split functions. Fully quaternion-valued activation functions are an extension to the hypercomplex domain of real-valued functions, such as sigmoid or hyperbolic tangent functions. Despite their better performance \cite{ujang2010quaternion}, careful training is needed due to the occurrence of singularities that can affect performance. To avoid this, split activation functions \cite{arena1994neural,ujang2010quaternion} have been presented as a simpler solution for QNNs. In split activation functions, a conventional real-valued function is applied component-wise to a quaternion, alleviating singularities while holding true the universal approximation theorem as demonstrated in \cite{arena1994neural}. Through their simplicity, split activation functions map quaternion space back to real-valued space by ignoring the nature of the relationship that exists between the components. This is captured in the following relationship, 

\begin{align}
\alpha((p_0, \vec{p})) = f(p_0) + f(p_1) i + f(p_2) j + f(p_3) k
\end{align}

where $\alpha$ corresponds to a split function and $\vec{p}=p_1 i + p_2 j + p_3 k$, and $f$ represents any standard and real-valued activation function such as Sigmoid, Tanh, and ReLU. In our experiments, we used ReLU activation.  Our network architecture shown in figure \ref{fig:qrbsa} consists of three parts: a shallow feature extraction module, a deep feature extraction module, and an upsampling and reconstruction module. 

\textbf{Shallow Feature Extractor:} 
This module uses a single quaternion convolution layer to reduce the spatial size of sparsely sectioned EBSD maps, while extracting shallow features. 
\begin{align}
    F_0 = H_{SF}(I_{LR})
\end{align}
Here, $I_{LR}$ is a sparsely sectioned 2D EBSD map and $H_{SF}(.)$ is a single quaternion convolution layer of kernel size $3 \times 3$, which has 4 input channels and 128 output channels. The generated shallow features ($F_0$) are given to the deep feature extractor module ($H_{DF})$. 

\textbf{Deep Feature Extractor:} 
To learn from sparsely sectioned EBSD maps, our deep feature extractor module uses stacked quaternion residual self-attention (QRSA) blocks to extract high-frequency information and long skip connection to bypass low-frequency information. Residual blocks allow for a deeper network architecture, which provides a larger receptive field and better training stability. In our QRSA module, we use both CNN and transformer ideas to combine the effectiveness of the locality of CNNs with the expressivity of transformers that enables them to synthesize high-resolution EBSD maps. The CNN structure offers local connectivity and translation equivariance, allowing transformer components to freely learn complex and long-range relationships.
Each quaternion residual self-attention (QRSA) block consists of two quaternion convolution layers and a piece-wise ReLU activation between them, and a quaternion transformer block. The quaternion convolution layers with piece-wise ReLU activation help in learning the local structure of extracted shallow features, while the quaternion transformer block captures long-range correlations among features. The short-skip connection is useful to bypass low-frequency information during training. 
\begin{gather}
    F_{DF} = H_{DF}(F_0) \\
    H_{DF} = QConv \circ QRSA_{n} \circ QRSA_{n-1} \circ ... QRSA_{i} ... \circ QRSA_0 + I  
\end{gather}
where, $H_{DF}(.)$ is a deep feature extractor module, and $F_{DF}$ is a 128 channels feature map which goes to the upscale and reconstruction module. $QConv$ is a quaternion convolution, $QRSA_{i}$ is a $i^{th}$ quaternion residual block, and I is a long skip connection. 

\textbf{Quaternion Transformer Block:} The standard transformer architecture \cite{vaswani2017attention} consists of a self-attention layer, a feedforward network, and layer normalization. The original transformer architectures \cite{vaswani2017attention, dosovitskiy2020image} are not suitable for restoration tasks due to the requirement of quadratic complexity of spatial size $\mathcal{O}(W^2 H^2)$, where W, H is the spatial size of images or EBSD maps. Similar to the approach of \cite{zamir2022restormer}, we compute attention maps across the features dimension, which reduces the problem to linear complexity. However instead of depthwise convolution, we use quaternion convolution, which can be considered as a combination of depthwise convolution and group convolution, but with four-dimensional quaternion constraints. We have also incorporated an equivalent quaternion-based gating mechanism into the feedforward network within the transformer,  and the traditional convolution used in \cite{zamir2022restormer} has been replaced with quaternion convolution layers to account for EBSD data modalities. Layer normalization plays a crucial role in the stability of training in transformer architectures. The quaternion layer-normalization is equivalent to the real-valued one, but adapted to quaternion algebra, and allows the building of deeper architectures by normalizing the output at each layer. From the normalized features, the quaternion self-attention layer first generates query (Q), key (K) and value (V) projections enriched with local context. After reshaping query and key projection to reshaped query ($Q_r$) and reshaped key ($K_r$), a transposed attention map (A) is generated. The refined feature map, which has global statistical information, is calculated from the dot product of the value projection ($V_r$) and the attention map (A).
\begin{align}
    \textrm{Attention (A)} = \textrm{Softmax}({K_r}. \frac{{Q_r}}{\alpha}) \\
    \textrm{Quat-SelfAttention}({Q_r}, {K_r}, {V_r}) = {V_r} \textrm{ . A} \\
    \textrm{Gating}(X_2) = \textrm{GeLU}(W_1(X_2)) \cirbd W_2(X_2) 
\end{align}
Where $\cirbd$ represents elementwise multiplication, and $W_{i}$ is a combination of quaternion convolution layers with kernel size 1 and 3, respectively. 

\begin{figure}
    \centering
    \includegraphics[width=1\linewidth]{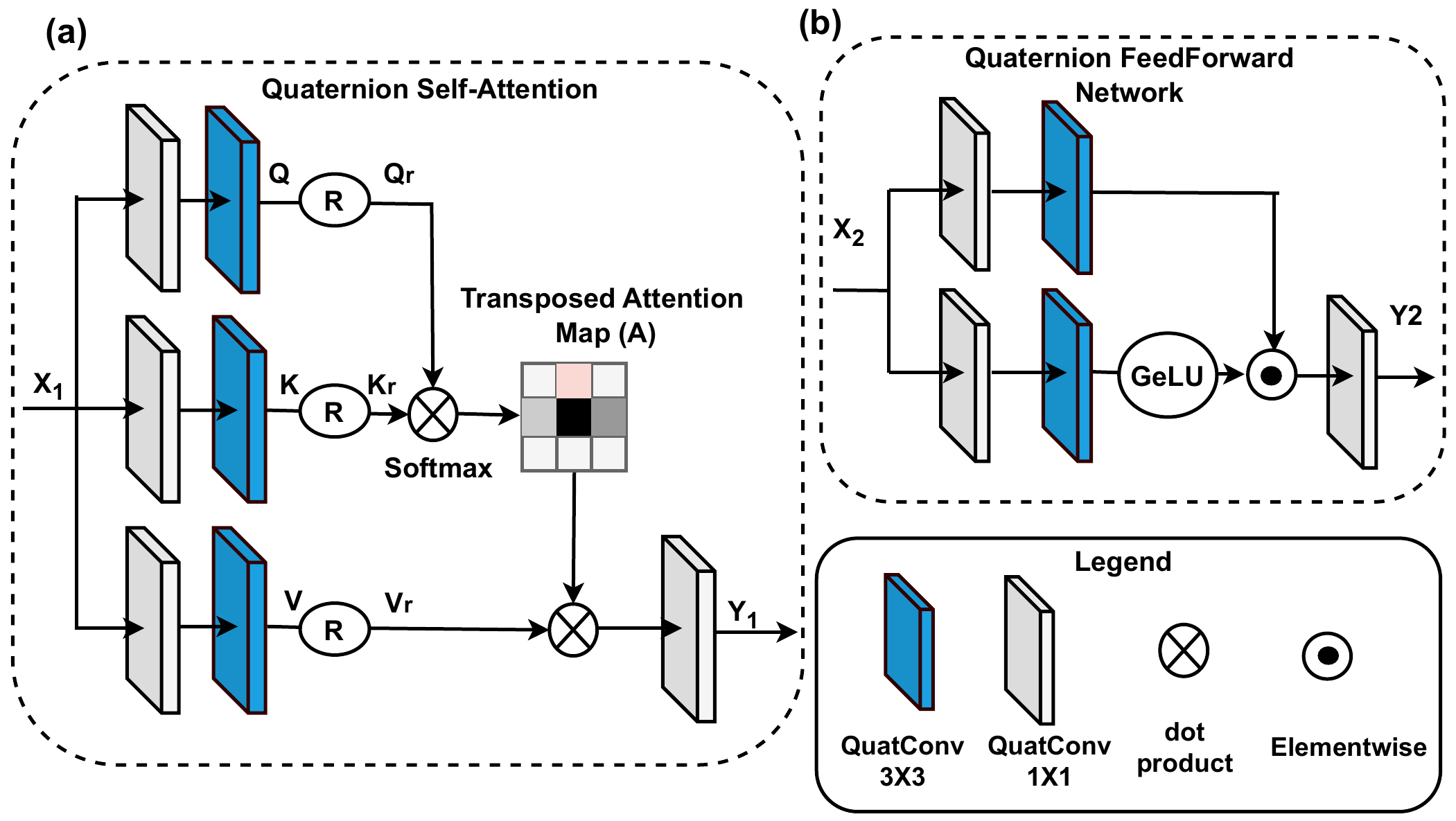}
    \caption{\textsc{Quaternion Self Attention:} Self-Attention in (a) is computed using quaternion convolution across feature dimension instead of spatial dimension to reduce computational complexity to linear. An attention map (A) is calculated from reshaped query ($Q_r$) and reshaped key ($K_r$). A refined feature map is computed from the attention map (A) and reshaped value ($V_r$). \textsc{Quaternion Feed Forward Network:} Shown in (b), performs controlled feature transformation to allow useful information to propagate further using gated quaternion convolution.}
    \label{fig:quat_attn}
\end{figure}

\textbf{Upsampling and Reconstruction:}
The upsampling and reconstruction module has a 1D pixelshuffle layer and a quaternion convolution layer of kernel size 3. The original pixelshuffle layer \cite{shi2016real} is designed for 2D upsampling, but we have modified it for 1D upsampling in our framework that generates information in z dimension. Each block of the upsampling and reconstruction module upsamples deep features by a factor of 2, with the number of blocks depending on the scaling factors. 
\begin{align}
    F_{\uparrow} = H_{\uparrow} (F_{DF})\\
    I_{SR} = H_R(F_{\uparrow}) 
\end{align}

Where, each block $H_{\uparrow}$ of the module has a 1D pixel-shuffle layer, and a quaternion convolution layer $H_R$ of kernel size $3 \times 3$.

\subsection{2D to 3D:}
The output of the QRBSA network is a 2D high-resolution EBSD map from a sparsely sectioned 2D EBSD map in the z direction. The 2D high-resolution EBSD maps are then combined to make a 3D volume. The missing z rows, as in figure \ref{fig:ebsd3d} (b), can be generated either from xz plane ($y_{normal}$) or yz plane ($x_{normal}$). Therefore, there are two ways to form the 3D volume. In this work, we generated both 3D volumes separately, but we plan to design an algorithm in the future to combine the xz plane and yz plane information to make a single 3D volume. 

\subsection{Qualitative Output Comparison:}
\begin{figure}
    \centering
    \includegraphics[width=1\linewidth]{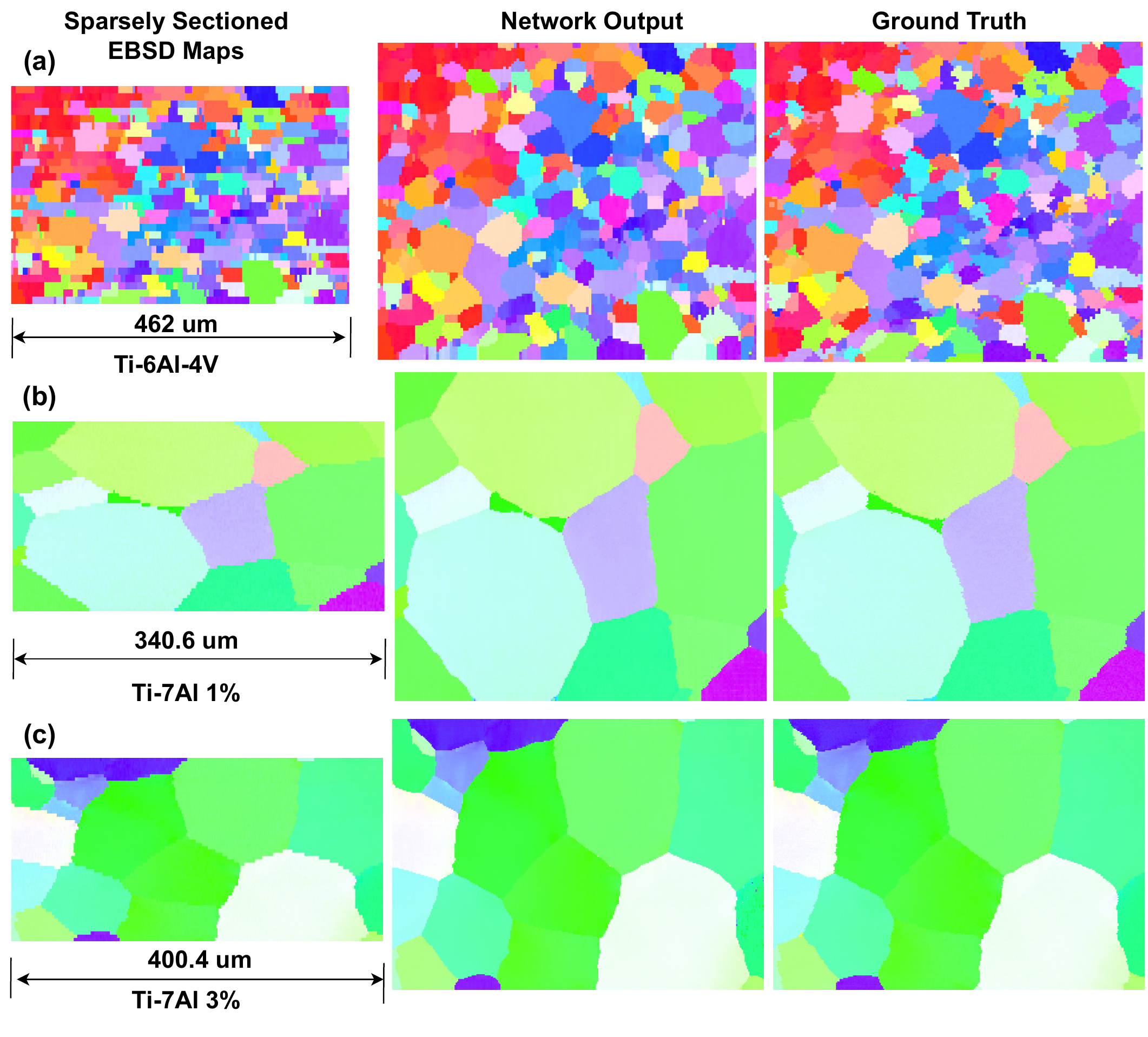}
    \caption{\textsc{Visual comparison of network output for example 2D EBSD maps with a scale factor of 4:} The predicted EBSD maps (Network Output) from the QRBSA network are similar to the ground truth EBSD maps in for both the Ti-6Al-4V dataset (a) and both Ti-7Al datasets (b) and (c). }
    \label{fig:qual_res_1d}
\end{figure}

\begin{figure}
    \centering
    \includegraphics[width=1\linewidth]{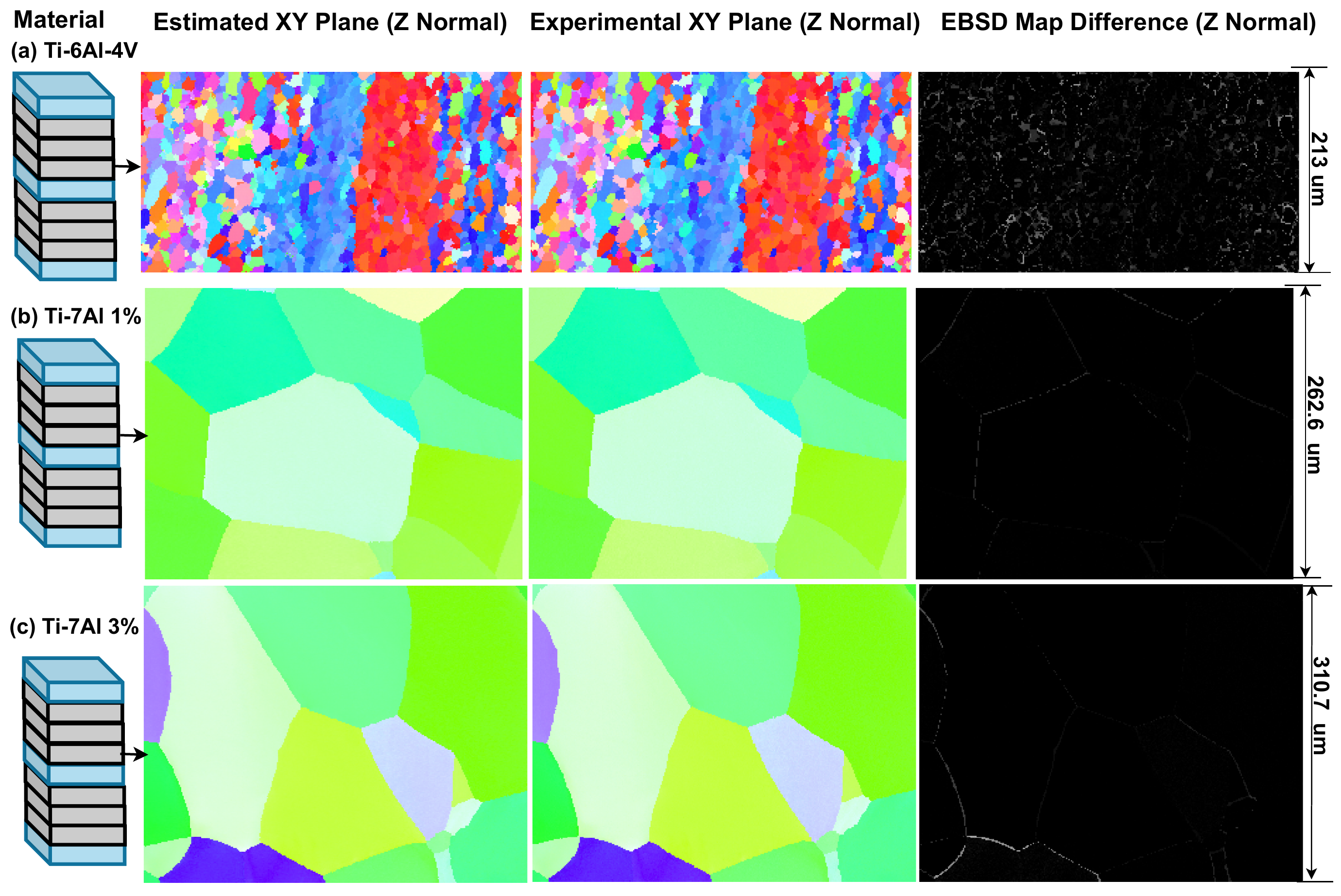}
    \caption{\textsc{Neural network output vs. ground truth with difference map:} The deep learning framework is able to estimate the missing xy planes due to sparse z-sampling (gray) with data that looks similar to the ground truth for Ti-6Al-4V in (a) and Ti-7Al in (b) and (c). The EBSD map difference column shows the intensity difference between ground truth and estimated EBSD maps, which indicate that learning grain shapes for Ti-6Al-4V is more difficult than for Ti-7Al, likely due to smaller grain size and more grain boundary regions.}
    \label{fig:qual_res_3d}
\end{figure}

The sparsely sectioned 3D EBSD data is downsampled by scale factors of 2, 4 in the z dimension by removing the xy planes ($z_{normal}$) to reflect how EBSD resolution would be reduced in a serial sectioning experiment. Our network QRBSA is trained on 2D orthogonal planes ($x_{normal}$ and $y_{normal}$) of paired sparsely sectioned EBSD maps and high-resolution EBSD maps, generating the high-resolution 2D maps in z dimension shown in figure \ref{fig:qual_res_1d}.  The most noticeable visual defects in 2D appear as pixel noise or short vertical lines, particularly around small grain features and high-aspect-ratio grains whose shortest axis is aligned with the z-direction.  In addition to planar output analysis, we can also create 3D volumes from the sparsely sectioned xz planar ($y_{normal}$) or yz planar ($x_{normal}$) EBSD maps, and then sample the xy planes ($z_{normal}$) from these volumes to evaluate how well the QRBSA is inferring missing z-sample planes, as shown in figure \ref{fig:qual_res_3d}. We can observe that our deep learning framework is able to completely predict omitted xy planes, comparably to the ground truth xy plane, with the exception of some shape variations around grain boundaries, particularly in Ti-6Al-4V. If the xy plane in figure \ref{fig:qual_res_3d} would have been omitted during experimental data collection, our framework would have estimated it with good accuracy.  
 
\subsection{Quantitative Output Comparison:}
\begin{table}[!ht]
\begin{tabular}{l|c|ll|ll|ll|}
Network & \multicolumn{1}{l|}{Trainable Parameters} & \multicolumn{2}{c|}{Ti-6Al-4V}                                                                                                                       & \multicolumn{2}{c|}{Ti-7Al 1\%}                                                                                                                        & \multicolumn{2}{c|}{Ti-7Al 3\%}                                                                                                                      \\ \hline
        & \multicolumn{1}{l|}{}                     & \multicolumn{1}{l|}{x2}                                                              & x4                                                            & \multicolumn{1}{l|}{x2}                                                              & x4                                                              & \multicolumn{1}{l|}{x2}                                                             & x4                                                             \\ \hline
HAN     & 63,315,578                                & \multicolumn{1}{l|}{\begin{tabular}[c]{@{}l@{}}21.21/\\ 0.839\end{tabular}}          & \begin{tabular}[c]{@{}l@{}}17.55/\\ 0.673\end{tabular}       & \multicolumn{1}{l|}{\begin{tabular}[c]{@{}l@{}}29.36/\\ 0.919\end{tabular}}        & \begin{tabular}[c]{@{}l@{}}26.54/\\ 0.849\end{tabular}         & \multicolumn{1}{l|}{\begin{tabular}[c]{@{}l@{}}31.26/\\ 0.941\end{tabular}}        & \begin{tabular}[c]{@{}l@{}}28.90/\\ 0.905\end{tabular}        \\ \hline
EDSR    & 6,355,460                                 & \multicolumn{1}{l|}{\begin{tabular}[c]{@{}l@{}}21.49/\\ 0.86\end{tabular}}           & \begin{tabular}[c]{@{}l@{}}18.09/\\ 0.718\end{tabular}        & \multicolumn{1}{l|}{\begin{tabular}[c]{@{}l@{}}30.06/\\ 0.944\end{tabular}}          & \begin{tabular}[c]{@{}l@{}}27.34/\\ 0.907\end{tabular}          & \multicolumn{1}{l|}{\begin{tabular}[c]{@{}l@{}}32.16/\\ 0.958\end{tabular}}         & \begin{tabular}[c]{@{}l@{}}29.57/\\ 0.932\end{tabular}         \\ \hline
QEDSR   & 1,593,092                                 & \multicolumn{1}{l|}{\begin{tabular}[c]{@{}l@{}}21.44/\\ 0.861\end{tabular}}          & \begin{tabular}[c]{@{}l@{}}18.04/\\ 0.710\end{tabular}        & \multicolumn{1}{l|}{\begin{tabular}[c]{@{}l@{}}29.89/\\ 0.935\end{tabular}}          & \begin{tabular}[c]{@{}l@{}}27.22/\\ 0.904\end{tabular}          & \multicolumn{1}{l|}{\begin{tabular}[c]{@{}l@{}}32.05/\\ 0.95\end{tabular}}          & \begin{tabular}[c]{@{}l@{}}29.52/\\ 0.93\end{tabular}          \\ \hline
QRBSA   & 5,952,782                                 & \multicolumn{1}{l|}{\textbf{\begin{tabular}[c]{@{}l@{}}21.60/\\ 0.870\end{tabular}}} & \textbf{\begin{tabular}[c]{@{}l@{}}18.2/\\ 0.730\end{tabular}} & \multicolumn{1}{l|}{\textbf{\begin{tabular}[c]{@{}l@{}}30.20/\\ 0.946\end{tabular}}} & \textbf{\begin{tabular}[c]{@{}l@{}}27.48/\\ 0.908\end{tabular}} & \multicolumn{1}{l|}{\textbf{\begin{tabular}[c]{@{}l@{}}32.36/\\ 0.96\end{tabular}}} & \textbf{\begin{tabular}[c]{@{}l@{}}29.65/\\ 0.94\end{tabular}} \\ \hline
\end{tabular}
 \caption{\textsc{\textbf{PSNR/SSIM:} Comparison of PSNR/SSIM and complexity for different networks for scale factors 2 and 4:}
  Columns represent number of trainable parameters and PSNR/SSIM for different titanium datasets. A larger number is desired for both PSNR/SSIM.}
 \label{tab:psnr_ssim_network_x4}
\end{table}

The image metrics of peak signal to noise ratio (PSNR) and structural similarity index measure (SSIM) were used to quantitatively evaluate the performance of the generated EBSD maps from the QRBSA network. We calculated the PSNR and SSIM of the Inverse Pole Figure (IPF) maps from the EBSD data for scale factors of 2 and 4 as shown in table \ref{tab:psnr_ssim_network_x4}. Higher PSNR and SSIM values represent more similarity with the ground truth. The PSNR/SSIM of Ti-6Al-4V is lower compared to Ti-7Al datasets due to its higher texture variability, wider range of orientations, and generally smaller grain features. We applied four different network architectures with different computational complexity to three different 3D EBSD datasets, as shown in table \ref{tab:psnr_ssim_network_x4}. Here, a simpler deep residual architecture (EDSR), outperforms a more complex holistic attention network (HAN) on EBSD data despite having significantly lower computational complexity. The amount of available EBSD data is significantly lower than open-source RGB image datasets, so simply increasing network complexity does not improve performance, as this added complexity demands additional training information and does not meaningfully consider relevant data modalities. QEDSR incorporates quaternion considerations in a similar architecture to EDSR, which greatly reduces in the number of network parameters, but also causes a slight drop in performance due to overall lack of complexity. We take advantage of this reduction in complexity to add in additional self-attention for better recognition of long-range patterns and global statistics. This QRBSA network demonstrates the best performance on EBSD map restoration, while still maintaining lower complexity than state-of-the-art residual architectures for single-image super-resolution tasks.

\section{Discussion}
Both quantitative and qualitative results demonstrate that this physics-based deep learning framework can accurately estimate the missing xy planes ($z_{normal}$) of 3D EBSD data for multiple variants of titanium alloys, both with a coarser polycrystalline structure (Ti-7Al) and finer structure with stronger texture (Ti-6Al-4V). In 2D inferred EBSD planes show noise around small features, mostly in the form of point and line defects in the z-direction associated with grains whose overall shape information was lost due to omission of sample planes in low resolution.  It is possible that a downsampling approach incorporating anti-aliasing could prevent this shape information loss \cite{Jung2021}, but this approach would not be reflective of actual experimental downsampling in 3D EBSD. This general shape loss effect, along with a larger number of small grains, varying local crystallographic texture, and a wide range of represented crystal orientations, made the Ti-6Al-4V the most difficult dataset for inference.  This is further evidenced by a larger number of grain boundary differences for Ti-6Al-4V in figure  \ref{fig:qual_res_3d}, as well as a lower PSNR/SSIM score in table \ref{tab:psnr_ssim_network_x4}.  Additional noise analysis for generated xy planes is shown in the supplement, and there is ongoing work to improve performance using 3D architectures and grain shape information \cite{jangid20223d} with adaptive multi-scale imaging in z dimension as more of this type of data becomes available. 

When considering data beyond the use-case sets tested, the approach shown here is directly applicable to any serial sectioning technique for gathering 3D EBSD information, including FIB sectioning, laser ablation, and robotic serial sectioning \cite{Chapman2021,RowenhorstCOSSMS2020}. Further, data from other 3D grain mapping techniques that rely on synchrotron X-ray sources such as diffraction contrast tomography (DCT) \cite{Reischig_COSSMS_2020} or high energy diffraction microscopy (HEDM) \cite{Bernier2020,Miller2020} may also be applicable for the infrastructure presented here. Similar approaches to this may be particularly useful in lab source DCT experiments \cite{Oddershede2019,Bachmann2019}, where the X-ray source constraints limit grain mapping resolution in comparison to synchrotron sources. For example, one could use difficult to acquire synchrotron X-ray mapping experiments as HR data to train a network to inform LR X-ray mapping experiments collected more routinely at the laboratory. 
 
In summary, we have designed a quaternion-convolution-based deep learning framework with crystallography physics-based loss to generate costly high-resolution 3D EBSD data from sparsely sectioned 3D EBSD data while accounting for the physical constraints of crystal orientation and symmetry. Alongside this, an efficient quaternion-based transformer block was developed to learn long-range trends and global statistics from EBSD maps.  Using quaternion convolution instead of regular convolution is critical for crystallographic data, both in terms of output quality and neural network complexity, as reducing the number of trainable parameters enables transformer addition without major complexity burden (see table \ref{tab:psnr_ssim_network_x4}). This framework can be directly applied to any experimental 3D EBSD approaches that rely on serial sectioning techniques to collect orientation information. 

\section{Methods}
\subsection{EBSD Datasets}
EBSD maps represent crystal orientations collected at each physical pixel location in crystalline materials, which are fundamentally anisotropic and atomically periodic. Orientations for each pixel within the network learning environment are expressed in terms of quaternions of the form $q = q_0 + iq_1 + jq_2 + kq_3$. The quaternions are suitable to design a physics-based loss function for deep learning framework \cite{dkjangid_ebsdsr}. To avoid redundancy in quaternion space (between $q$ and $-q$), all orientations are expressed with their scalar $q_0$ as positive. For visualization according to established conventions, quaternions are reduced to the Rodrigues space fundamental zone based on space group symmetry, converted into Euler angles, and projected using inverse pole figure (IPF) projection using the open-source Dream3D software \cite{groeber2014dream}, as shown in our previous work \cite{dkjangid_ebsdsr}. Ground truth 3D EBSD datasets were experimentally collected from titanium alloy samples: Ti-6Al-4V and Ti-7Al (one Ti-7Al sample deformed in tension to 1\%, and one to 3\%), using a commercially-available rapid-serial-sectioning electron microscope referred to as the TriBeam \cite{Echlin2015MatChar,Echlin2021}. sparsely sectioned EBSD datasets are created by removing xy ($z_{normal}$) planes from the high-resolution ground truth with a downscale factor of 2 and 4 (LR = $\frac{1}{4}$HR or LR=$\frac{1}{2}$HR). This is done to imitate the beam raster steps that would occur in a Tribeam experiment with more sparsely sectioned EBSD data, which would not influence the electron beam-material interaction volume at each location, but rather lead to greater raster spacing step sizes. More information about dataset pre-processing are given in the supplementary material.  
 
\subsection{Network Implementation and Output Evaluation}
We use a learning rate of 0.0002, an Adam optimizer with $\beta_1$ = 0.9, $\beta_2$=0.99, ReLU activation, batch size of 4 and downscaling factor of 2 and 4. The patch size of HR EBSD maps is selected from [16, 32, 64, 100] during training of the network. The framework is implemented in PyTorch and trained on NVIDIA Tesla V100 GPU for 2000 epochs, which took approximately 100 hours. Once training is completed, inference time for a given 2D LR EBSD map is on the order of less than one second for an imaging area that would normally take about 10 minutes to gather manually.

\subsection{Loss:}
The QRBSA network is trained using a physics based loss function \cite{dkjangid_ebsdsr}, which uses rotational distance approximation loss with enforced hexagonal crystal symmetry (HCP). Rotational distance loss computes the misorientation angles between the predicted and ground truth EBSD map in the same manner that they would be measured during crystallographic analysis, with approximations to avoid discontinuities at the edge of the fundamental zone. The rotational distance between two quaternions can be computed as the following:

\begin{align}
     \theta &= 4 \sin^{-1} \left ( \frac{d_{\text{euclid}}}{2} \right )
\end{align}

where, $d_{\text{euclid}} = \norm{q_1 - q_2}_2$. While $d_{\text{euclid}}$ is Lipschitz, the gradient of $\theta$ goes to $\infty$ as $d_{\text{euclid}} \rightarrow 2$. To address this issue during neural network training, a linear approximation was computed at $d_{\text{euclid}} = 1.9$, and utilized for points $> 1.9$. 

\subsection{Progressive Learning:}
In our previous work \cite{dkjangid_ebsdsr}, we used a fixed patch size of dimension $64 \times 64$ for training the CNN based architectures which help in learning local correlations. However, self-attention is required to have larger patch sizes, which aids in learning global correlations. Inspired from the work of Zamir \cite{zamir2022restormer}, we use progressive patch samples from sizes of [16, 32, 64, 100] in the training process to learn global statistics. We start from a smaller patch size in early epochs and increase to a larger patch sizes in the later epochs. The progressive learning acts like the curriculum learning process where a network starts with a simple tasks and gradually moves to learning a more complex one. 

\section{Data Availability}
Pre-trained versions of the network modules produced in this paper will be made publicly available through the BisQue cyberinfrastructure at {\color{blue}https://bisque2.ece.ucsb.edu}. Material datasets will be available by request at the discretion of the authors. 

\section{Code Availability}
Architecture code will be made publicly accessible through GitHub.

\bibliography{source}

\section{Acknowledgements}
This research is supported in part by NSF award number 1664172. N.Brodnik gratefully acknowledges financial support from NSWC Grant (N00174-22-1-0020). The authors gratefully acknowledge Patrick Callahan, Toby Francis, Andrew Polonsky, and Joseph Wendorf for collection of the 3D Ti-6Al-4V and Ti-7Al datasets. The MRL Shared Experimental Facilities are supported by the MRSEC Program of the NSF under Award No. DMR 1720256; a member of the NSF-funded Materials Research Facilities Network (www.mrfn.org). Use was also made of computational facilities purchased with funds from the National Science Foundation (CNS-1725797) and administered by the Center for Scientific Computing (CSC). The CSC is supported by the California NanoSystems Institute and the Materials Research Science and Engineering Center (MRSEC; NSF DMR 1720256) at UC Santa Barbara. Use was made of the computational facilities purchased with funds from the National Science Foundation CC* Compute grant (OAC-1925717) and administered by the Center for Scientific Computing (CSC). The ONR Grant N00014-19-2129 is also acknowledged for the titanium datasets. The authors of this work declare no competing financial or non-financial interests.

\section{Author Contributions}

\begin{itemize}
    \item D.K. Jangid: Development of complete framework, Design and coding of complete network architecture, Preprocessing of datasets for training the network, All experiments to make our idea work, Qualitative and quantitative evaluation (image metrics) of results, Manuscript Preparation 
    \item N.R. Brodnik: Initial preparation of training, validation, and test datasets , Feedback on generated results, Mentoring, Conception of Ideas, Manuscript Preparation 
    \item M.P. Echlin: Preparation of datasets, Conception of ideas, Mentoring,  Feedback on generated results from network architecture, Manuscript Preparation 
    \item S.H. Daly (Co-PI): Conception of ideas, Mentoring, Funding Acquisition, Manuscript Preparation 
    \item T.M. Pollock (Co-PI): Conception of ideas, Mentoring, Funding Acquisition, Manuscript Preparation 
    \item B.S. Manjunath (Co-PI): Conception of ideas, Mentoring, Funding Acquisition, Manuscript Preparation 
    \item All authors contributed to the writing of this manuscript with writing efforts being led by D.K. Jangid, N.R. Brodnik and M.P. Echlin.
\end{itemize}

\section{Competing Interests}
The Authors declare that there are no competing interests.

\section{Supplementary Material}

\subsection{Quaternion Convolution Neural Network:}
A quaternion q is a four component number of the form $q = q_0 + iq_1 + jq_2 + kq_3$, where basis vectors $(i,j,k)$ satisfy the following relationship:
\begin{align}
     i^2 = j^2 = k^2 = ijk = -1
\end{align}

The quaternion can be represented as matrix of real numbers.

\begin{align}
 \begin{bmatrix}
q_0 & q_1 & q_2 & q_3\\
-q_1 & q_0 & -q_3 & q_2 \\
-q_2 & q_3 & q_0 & -q_1 \\
-q_3 & -q_2 & q_1 & q_0
\end{bmatrix}   
\end{align}
In a Quaternion Convolution Neural Network (QCNN), the convolution of a quaternion filter matrix with a quaternion vector is Hamilton product. The Hamilton product can also be represented as a matrix of real numbers.

\begin{align}
    p \otimes q = 
    \begin{bmatrix}
    p_0 & -p_1 & -p_2 & -p_3 \\
    p_1 & p_0 & -p_3 & p_2 \\
    p_2 & p_3 & p_0 & -p_1 \\
    p_3 & -p_2 & p_1 & p_0
    \end{bmatrix}
    *
    \begin{bmatrix}
    q_0 \\
    q_1 \\
    q_2 \\
    q_3
    \end{bmatrix}
\end{align}

\begin{figure}
    \centering
    \includegraphics[width=1\linewidth]{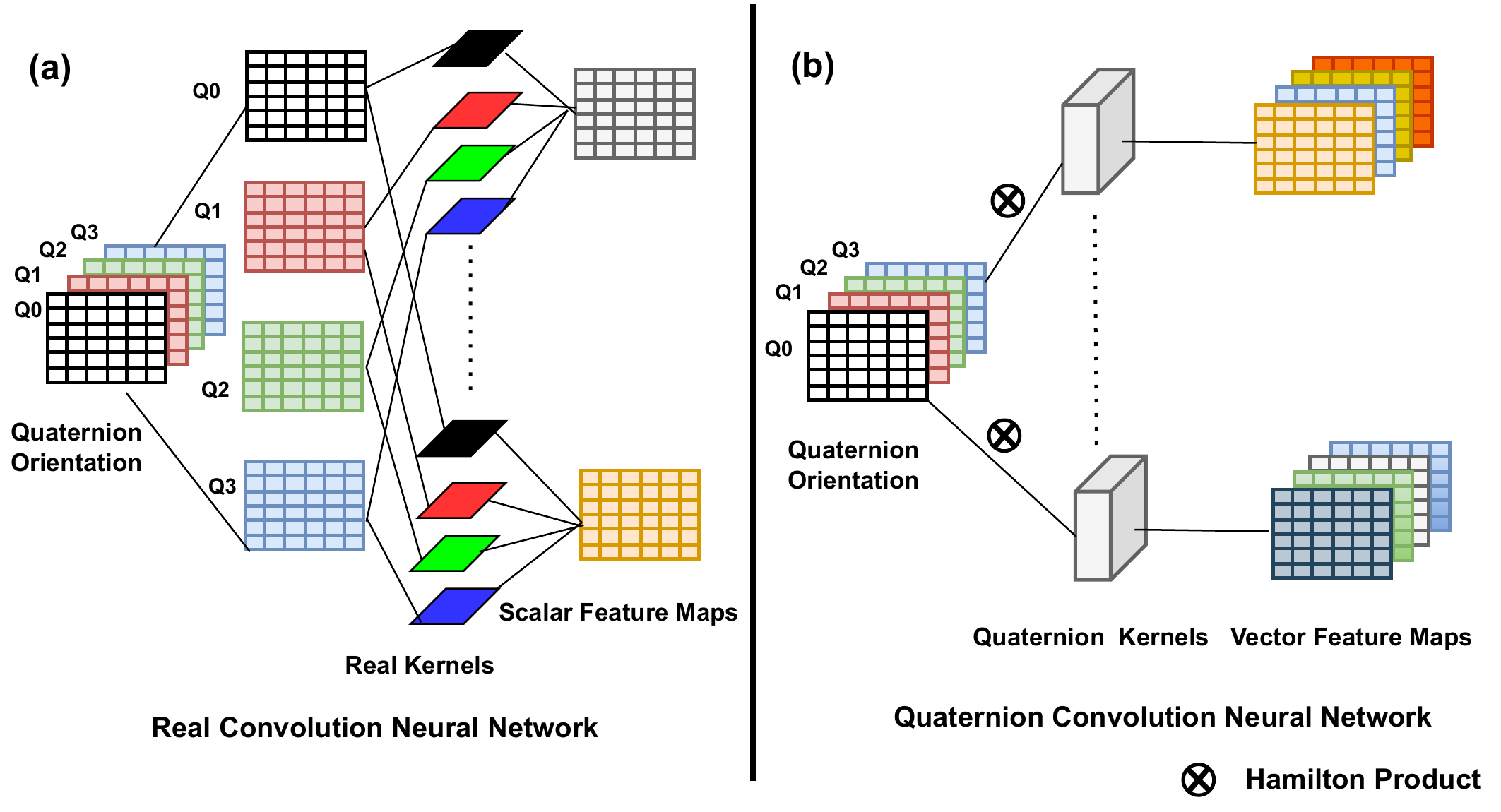}
    \caption{\textsc{Quaternion Convolution:} In a real convolution neural network, correlation, a scalar quantity, is computed between independent kernels and channel dimension of quaternion feature maps. In a quaternion convolution neural network, the Hamilton product, a vector quantity, is computed between quaternion kernels and quaternion feature input maps. The parameters in quaternion kernel is restricted according to quaternion algebra \cite{parcollet2020survey}.}
    \label{fig:qconv_op}
 \end{figure}

The difference between real convolution neural network (RCNN) and quaternion convolution neural network (QCNN) lies on how a basic convolution operation is performed as shown in figure \ref{fig:qconv_op}.  In RCNN, scalar feature maps are computed by computing correlations between channels of feature maps and kernel filters independently without considering inter-channel relationship. But, vector feature maps are generated using the Hamilton product between feature maps and quaternion kernels in QCNN. Both quaternion kernels and Hamilton products take inter-channel relationships into consideration.   

\subsection{Data Preprocessing}

\begin{figure}
    \centering
    \includegraphics[width=0.7
    \linewidth, height=6cm]{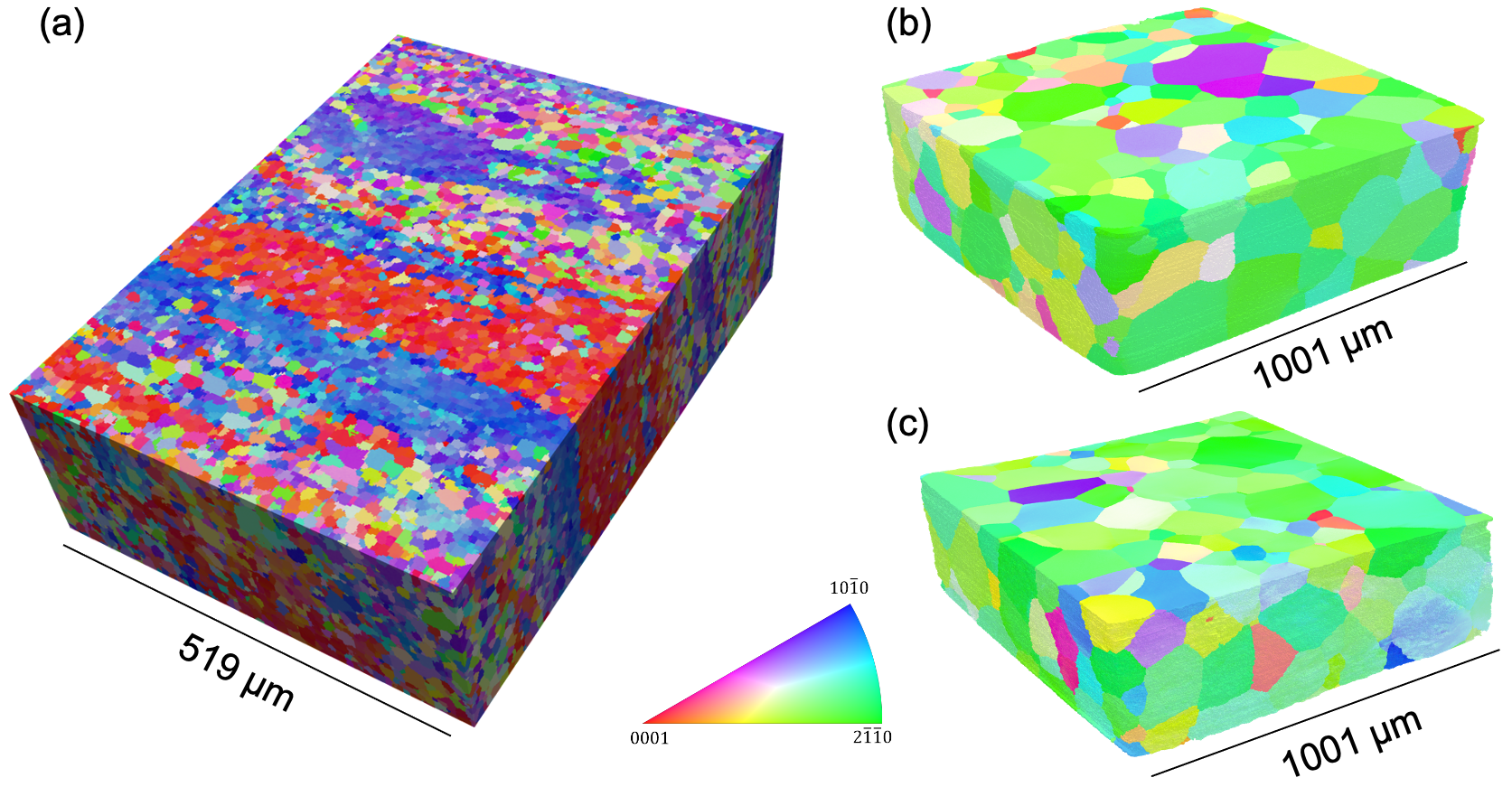}
    \caption{\textsc{Rendering of 3D EBSD dataset Investigated:} shown in IPF coloring of the titanium alloys, (a) Ti-6Al-4V and (b) Ti-7Al mechanically loaded in tension to 1\% strain and (c) 3\%, used for network training, testing and validation. Dataset details are available elsewhere \cite{Hemery2019,Charpagne2021}.}
    \label{fig:titanium_data}
\end{figure}

\textbf{High-Resolution (HR) Ground Truth:} 
The ground truth data is experimental 3D EBSD data gathered from the titanium alloys, Ti-6Al-4V and Ti-7Al (one Ti-7Al sample deformed in tension to 1\% and one to 3\%), using a commercially-available rapid-serial-sectioning electron microscope known as the Tribeam \cite{Echlin2015MatChar,Echlin2020COSSMS}. The Ti-6Al-4V dataset, shown in figure \ref{fig:titanium_data}(a), is of pixel size $346 \times 142 \times 471 \times 4$ ($z \times y \times x \times ch$), where the last dimension is the quaternion component. Analogously, the Ti-7Al shown in \ref{fig:titanium_data}(b) and \ref{fig:titanium_data}(c) are of size $232 \times 674 \times 770 \times 4$ ($z \times y \times x \times ch$) and $224 \times 770 \times 770 \times 4$ ($z \times y \times x \times ch$) pixels respectively, with all edges cropped to produce a perfect parallelpiped volume. Each voxel in the Ti-6Al-4V set has resolution of $1.5 \times 1.5 \times 1.5\, \mu$m, and in both Ti-7Al sets, each voxel has a resolution of $1.3 \times 1.3 \times 1.3\, \mu$m. 
 
These titanium alloys are composed primarily of the hexagonal close packed grains. In total, the Ti-6Al-4V dataset contains about 57,000 grains, visible in the IPF maps as regions of different color. The Ti-7Al material has larger grain size, with 500-1000 grains in each dataset. The datasets were proportionally divided into training, validation, and test subsets in ratios of 65\%, 15\%, and 20\% respectively. 
\vspace{10pt}

\noindent \textbf{Sparsely Sectioned Input EBSD Data:}
Sparsely sectioned EBSD data are downscaled versions of the high-resolution EBSD data.  However, because of how EBSD information is gathered, these EBSD data are not downscaled using pixel averaging.  Instead sparsely sectioned EBSD data are produced by removal of xy planes in the z direction from the high-resolution ground truth with a downscale factor of 4x and 2x (LR = $\frac{1}{4}$HR or LR=$\frac{1}{2}$HR). This is done to imitate the beam raster steps that would occur in an EBSD experiment with sparsely sectioned EBSD data, which would not influence the electron beam-material interaction volume at each location, but rather lead to greater raster distance between consecutive locations. 
 
 \begin{figure}
     \centering
     \includegraphics[width=1\linewidth]{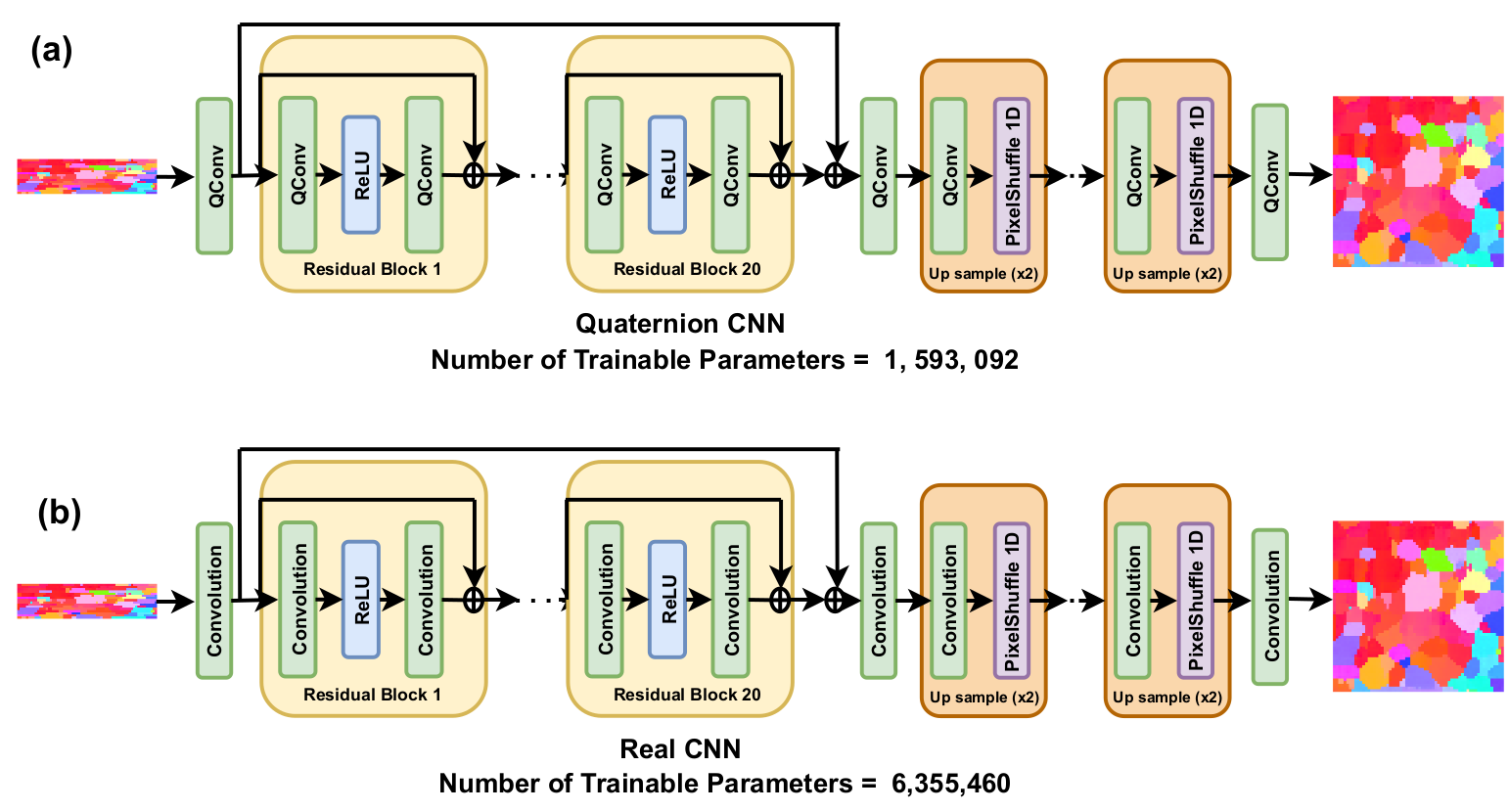}
      \caption{\textsc{Quaternion CNN and Real CNN:} Both QCNN and RCNN have same number of layers except basic convolution operation layer. In QCNN, total number of trainable parameters are reduced significantly which give us room to add more complexity such self-attention layer to learn global features.}
     \label{fig:qcnn_rcnn}
 \end{figure}

\subsection{Ablation Study}
To show the effectiveness of quaternion convolution, we have done experiments with an enhanced deep scale residual (EDSR) network \cite{lim2017edsr} and a quaternion enhanced deep residual (QEDSR) network as shown in figure \ref{fig:qcnn_rcnn}. The QEDSR is similar to EDSR except that the real-convolution layer is replaced by quaternion convolution layer that helps in reducing computational complexity. With decreased complexity due to the quaternion convolution layer, we can add a quaternion transformer block into our network architecture to learn global features of EBSD maps. The PSNR/SSIM values for the generated EBSD maps and number of trainable parameters of different neural network architectures are shown in table \ref{tab:psnr_ssim_network_x2}. Our network QRBSA is giving better PSNR and SSIM than EDSR \cite{lim2017edsr}, while having fewer trainable parameters. To understand the effect of the quaternion transformer block in QRBSA, we have trained our network with and without the quaternion transformer block as shown in table \ref{tab:psnr_ssim_sa}. We see an improvement in PSNR and SSIM with the quaternion transformer block for the Ti6-Al-4V dataset, which has more crystallographic texturing and is difficult to learn because of the mismatched lengthscales over which it persists, compared to the grain structure. The quaternion transformer block will become even more useful for learning global features as we increase our library of 3D datasets available to our deep learning framework in future.

\begin{figure}
    \centering
    \includegraphics[width=1\linewidth]{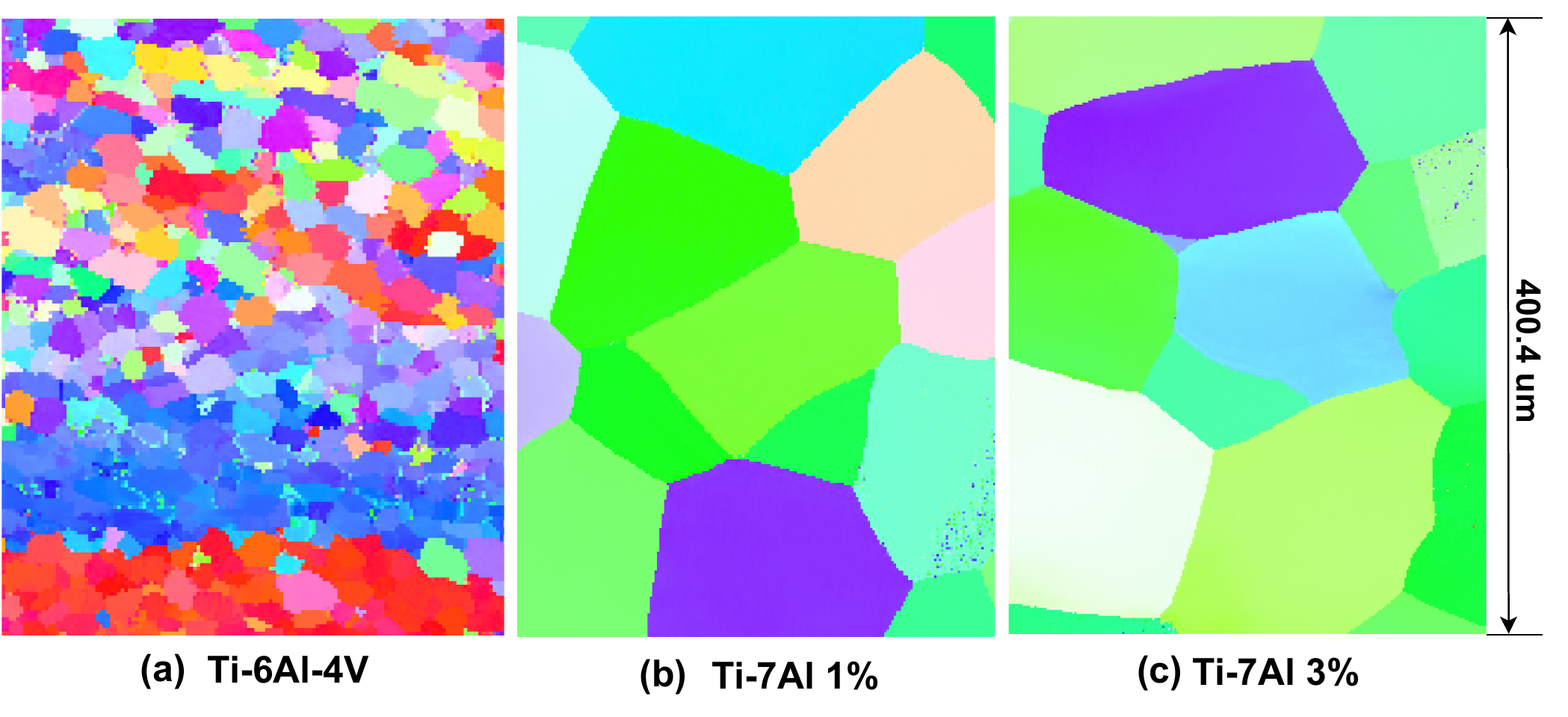}
    \caption{\textsc{Noise in Estimated XY Plane:} The deep learning framework is able to estimate the missing xy planes in z dimension but there are some minor noises in some of the xy planes.}
    \label{fig:noise_qual_res_3d}
\end{figure}

During analysis of our results, we found that some of generated xy planes ($z_{normal}$) in all three of the investigated datasets (Ti-6Al-4V, Ti-7Al 1 \%, and Ti-7Al 3 \%) have minor noise, as shown in \ref{fig:noise_qual_res_3d}. In our future work, we will be focusing on reducing this noise in the estimated xy planes by using 3D neural network architecture.  

\begin{table}[!ht]
\centering
\begin{tabular}{l|l|l|l|l|}
                                    Network &                               Trainable Parameters & 
                                    Ti-6Al-4V & 
                                    Ti-7Al 1\% & 
                                    Ti-7Al 3\%                                                                                                              \\ \hline
 \begin{tabular}[c]{@{}l@{}}HAN\end{tabular}                 &
\begin{tabular}[c]{@{}l@{}} 63,315,578 \end{tabular}                 &

\begin{tabular}[c]{@{}l@{}}21.21/\\0.839 \end{tabular} & 
\begin{tabular}[c]{@{}l@{}}29.36/\\0.919 \end{tabular} & 
\begin{tabular}[c]{@{}l@{}}31.26/\\0.941 \end{tabular}          \\ \hline 

\begin{tabular}[c]{@{}l@{}}EDSR \end{tabular}                 &
\begin{tabular}[c]{@{}l@{}}6,355,460 \end{tabular}                 &

\begin{tabular}[c]{@{}l@{}}21.49/\\0.86 \end{tabular} & 
\begin{tabular}[c]{@{}l@{}}30.06/\\0.944 \end{tabular} & 
\begin{tabular}[c]{@{}l@{}}32.16/\\0.958 \end{tabular}          \\ \hline

\begin{tabular}[c]{@{}l@{}}QEDSR \end{tabular}               &
\begin{tabular}[c]{@{}l@{}}1,593,092 \end{tabular}                 &
\begin{tabular}[c]{@{}l@{}}21.44/\\ 0.861\end{tabular} & 
\begin{tabular}[c]{@{}l@{}}29.89/\\ 0.935 \end{tabular}          & 
\begin{tabular}[c]{@{}l@{}}32.05/\\ 0.95\end{tabular}  \\ \hline

\begin{tabular}[c]{@{}l@{}}QRBSA \end{tabular}               &
\begin{tabular}[c]{@{}l@{}}5,952,782 \end{tabular}                 &
\textbf{\begin{tabular}[c]{@{}l@{}}21.60/\\ 0.870\end{tabular}} & 
\textbf{\begin{tabular}[c]{@{}l@{}}30.20/\\ 0.946\end{tabular}}          & 
\textbf{\begin{tabular}[c]{@{}l@{}}32.36/\\ 0.96\end{tabular}}  \\ \hline

\end{tabular}
\caption{\textsc{\textbf{PSNR/SSIM:} Comparison of PSNR/SSIM and Complexity for different networks for scale factor 2:}
 Columns represent number of trainable parameters and PSNR/SSIM for different titanium datasets. A larger number is desired for both PSNR/SSIM.}
\label{tab:psnr_ssim_network_x2}
\end{table}

\begin{table}[!ht]
\centering
\begin{tabular}{l|l|l|l|}
                                    Experiments &                    
                                    Ti-6Al-4V & 
                                    Ti-7Al 1\% & 
                                    Ti-7Al 3\%                                                                                                              \\ \hline
\begin{tabular}[c]{@{}l@{}}QRBSA without \\ quat transformer \end{tabular}                 &
\begin{tabular}[c]{@{}l@{}}18.12/\\0.718 \end{tabular} & 
\begin{tabular}[c]{@{}l@{}}27.27/\\0.906 \end{tabular} & 
\begin{tabular}[c]{@{}l@{}}29.49/\\0.9314 \end{tabular}          \\ \hline

\begin{tabular}[c]{@{}l@{}}QRBSA with  \\ quat transformer \end{tabular}               &
\textbf{\begin{tabular}[c]{@{}l@{}} 18.2/\\ 0.73\end{tabular}} & 
\textbf{\begin{tabular}[c]{@{}l@{}} 27.48/\\ 0.908 \end{tabular}}          & 
\textbf{\begin{tabular}[c]{@{}l@{}} 29.65/\\ 0.94\end{tabular}}  \\ \hline

\end{tabular}
\caption{\textsc{\textbf{PSNR/SSIM:} Comparison of PSNR/SSIM with and without quaternion transformer for scale factor 4:}
 Columns represent PSNR/SSIM for different titanium datasets. A larger number is desired for both PSNR/SSIM.}
\label{tab:psnr_ssim_sa}
\end{table}

\end{document}